\documentclass[10pt,twocolumn,letterpaper]{article}

\usepackage{cvpr}              
\usepackage{graphicx}
\usepackage{url}            
\usepackage{booktabs}       
\usepackage{amsfonts}       
\usepackage{nicefrac}       
\usepackage{microtype}      
\usepackage{xcolor}         

\usepackage{pifont}
\usepackage{color}
\usepackage{amssymb}
\usepackage{wrapfig}
\usepackage{makecell}
\usepackage{colortbl}
\usepackage{multirow}
\usepackage{fixltx2e}
\usepackage{subcaption}
\usepackage{amsmath}
\usepackage{etoolbox}
\usepackage{multicol}
\usepackage{refcount}
\usepackage{enumitem}
\usepackage[misc]{ifsym}

\definecolor{LightCyan}{rgb}{0.88,1,1}
\definecolor{mygray}{gray}{0.9}
\definecolor{mygray2}{gray}{0.6}
\usepackage[accsupp]{axessibility}  

\usepackage[pagebackref,breaklinks,colorlinks]{hyperref}

\hypersetup{
    colorlinks=true,
    linkcolor=blue,
    filecolor=magenta,      
    urlcolor=cyan,
    pdftitle={MixMAE},
    pdfpagemode=FullScreen,
    }

\usepackage[capitalize]{cleveref}
\crefname{section}{Sec.}{Secs.}
\Crefname{section}{Section}{Sections}
\Crefname{table}{Table}{Tables}
\crefname{table}{Tab.}{Tabs.}

\begin{document}

\title{MixMAE: Mixed and Masked Autoencoder for Efficient Pretraining of Hierarchical Vision Transformers}

\author{%
  Jihao Liu$^{1,2}$ \quad Xin Huang$^{2}$ \quad Jinliang Zheng$^{2}$ \quad Yu Liu$^{2}$~\textsuperscript{\Letter} \quad Hongsheng Li$^{1,3}$ \\
  $^1$ CUHK MMLab \\
  $^2$ SenseTime Research \\
  $^3$ CPII under InnoHK
}

\maketitle

\makeatletter{\renewcommand*{\@makefnmark}{}
\footnotetext{\textsuperscript{\Letter} Corresponding author.}\makeatother}

\begin{abstract}
   In this paper, we propose Mixed and Masked AutoEncoder (MixMAE), a simple but efficient pretraining method that is applicable to various hierarchical Vision Transformers. Existing masked image modeling (MIM) methods for hierarchical Vision Transformers replace a random subset of input tokens with a special $\mathrm{[MASK]}$ symbol and aim at reconstructing original image tokens from the corrupted image. However, we find that using the $\mathrm{[MASK]}$ symbol greatly slows down the training and causes pretraining-finetuning inconsistency, due to the large masking ratio (e.g., 60\% in SimMIM). On the other hand, MAE does not introduce $\mathrm{[MASK]}$ tokens at its encoder at all but is not applicable for hierarchical Vision Transformers.
   To solve the issue and accelerate the pretraining of hierarchical models, we replace the masked tokens of one image with visible tokens of another image, i.e., creating a mixed image. We then conduct dual reconstruction to reconstruct the two original images from the mixed input, which significantly improves efficiency.
    While MixMAE can be applied to various hierarchical Transformers, this paper explores using Swin Transformer with a large window size and scales up to huge model size (to reach 600M parameters).
    Empirical results demonstrate that MixMAE can learn high-quality visual representations efficiently. Notably, MixMAE with Swin-B/W14 achieves 85.1\% top-1 accuracy on ImageNet-1K by pretraining for 600 epochs.
    Besides, its transfer performances on the other 6 datasets show that MixMAE has better FLOPs / performance tradeoff than previous popular MIM methods.
    Code is available at \url{https://github.com/Sense-X/MixMIM}.
\end{abstract}

\section{Introduction}

Utilizing unlabeled visual data in self-supervised manners to learn representations is intriguing but challenging.
Following BERT~\cite{bert} in natural language processing, pretraining with masked image modeling (MIM) shows great success in learning visual representations for various downstream vision tasks~\cite{mae,beit,simmim,MaskedVP,videomae}, including image classification~\cite{imagenet}, object detection~\cite{coco}, semantic segmentation~\cite{ade20k}, video classification~\cite{somethingsomething}, and motor control~\cite{MaskedVP}. While those state-of-the-art methods~\cite{beit,mae} achieved superior performance on vanilla Vision Transformer (ViT)~\cite{vit,vaswani2017attention}, it is still an open question that how to effectively pretrain hierarchical ViT to purchase further efficiencies~\cite{swin,pvt,chu2021twins,liu2022uninet,dai2021coatnet} on broad vision tasks.

In general, existing MIM approaches replace a portion of input tokens with a special $\mathrm{[MASK]}$ symbol and aim at recovering the original image patches~\cite{beit,simmim}. However, using $\mathrm{[MASK]}$ symbol leads to two problems. On the one hand, the $\mathrm{[MASK]}$ symbol used in pretraining never appears in the finetuning stage, resulting in pretraining-finetuning inconsistency~\cite{bert}. On the other hand, the pretrained networks waste much computation on processing the less informative $\mathrm{[MASK]}$ symbols, making the pretraining process inefficient. 
Those problems become severer when a large masking ratio is used~\cite{mae,simmim,beit,videomae}. For example, in SimMIM~\cite{simmim}, a masking ratio of 60\% is used during the pretraining, i.e., 60\% of the input tokens are replaced with the $\mathrm{[MASK]}$ symbols. As a result, SimMIM needs relatively more epochs (i.e., 800) for pretraining. In addition, as the high masking ratio causes much pretraining-finetuning inconsistency, the performances of SimMIM on downstream tasks are limited.

In contrast, MAE~\cite{mae} does not suffer from the above problems by discarding the masked tokens in the encoder and uses the $\mathrm{[MASK]}$ symbols only in the lightweight decoder. MAE utilizes the vanilla ViT~\cite{vit} as the encoder, which can process the partial input efficiently with the self-attention operation. However, the design also limits the application of MAE on hierarchical ViTs as the hierarchical ViTs cannot process 1D token sequences with arbitrary lengths~\cite{swin,pvt}.

\begin{figure*}
    \centering
    \includegraphics[width=0.9\linewidth]{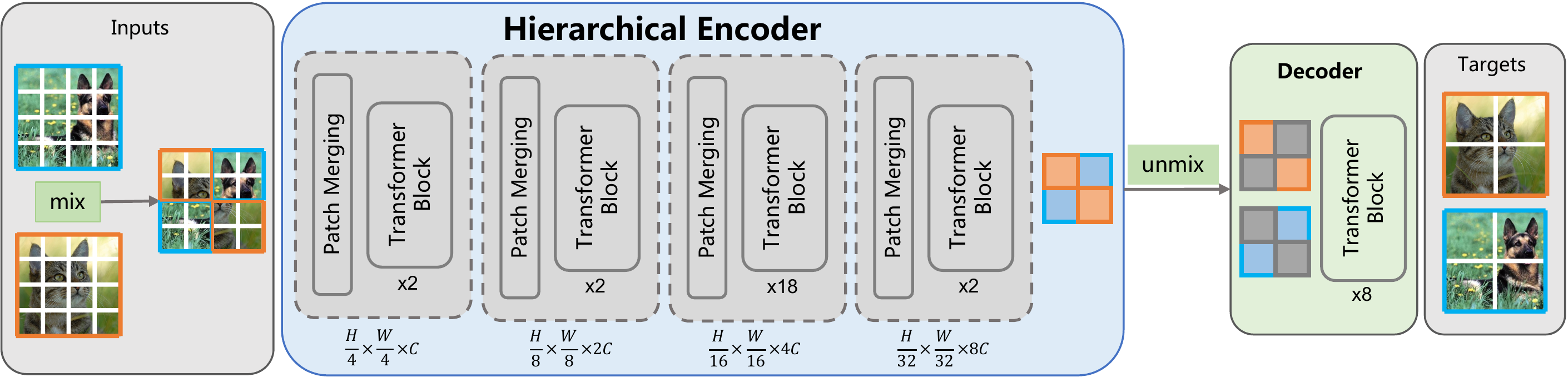}
    \vspace{-1em}
    \caption{
    Overview of MixMAE. For pretraining, two images are mixed with a random mixing mask to create a mixed image. MixMAE takes the mixed image as input and reconstructs the two original images. Right before decoding, the token embeddings are unmixed and filled with mask tokens for dual reconstruction of the two original images.}
    \label{fig:xmnet}
\end{figure*}

In this work, we propose MixMAE, a generalized pretraining method that takes advantage of both SimMIM~\cite{simmim} and MAE~\cite{simmim} while avoiding their limitations. 
In particular, given two random images from the training set, MixMAE creates a mixed image with random mixing masks as input and trains a hierarchical ViT to reconstruct the two original images to learn visual representations. 
From one image's perspective, instead of replacing the masked tokens of the image with the special $\mathrm{[MASK]}$ symbols, the masked tokens are replaced by visible tokens of the other image.
MixMAE adopts an encoder-decoder design. The encoder is a hierarchical ViT and processes the mixed image to obtain hidden representations of the two partially masked images. Before the decoding, the hidden representations are unmixed and filled with the $\mathrm{[MASK]}$ tokens. Following MAE~\cite{mae}, the decoder is a small ViT to reconstruct the two original images.
We illustrate the proposed MixMAE in Figure~\ref{fig:xmnet}.

\begin{table}[t]
    \centering
    \resizebox{1.0\linewidth}{!}{
    \begin{tabular}{lccc}
        \toprule
        Approach & \makecell{Compatible with \\ hierarchical ViT} & \makecell{Pretraining \\ efficient} & \makecell{Pretrain-finetune \\ consistent} \\
        \midrule
        BEiT \cite{beit} & \ding{51} & \textcolor{mygray2}{\ding{55}} & \textcolor{mygray2}{\ding{55}} \\
        SimMIM \cite{simmim} & \ding{51} & \textcolor{mygray2}{\ding{55}} & \textcolor{mygray2}{\ding{55}} \\
        MAE \cite{mae} & \textcolor{mygray2}{\ding{55}} & \ding{51} & \ding{51} \\
        MixMAE & \ding{51} & \ding{51} & \ding{51} \\
        \bottomrule
    \end{tabular}
    }
    \vspace{-1em}
    \caption{Key differences between MixMAE and related works.}
    \label{tab:difference}
\end{table}

MixMAE can be widely applied to pretrain different hierarchical ViTs, such as Swin Transformer~\cite{swin}, Twins~\cite{chu2021twins}, PVT~\cite{pvt}, etc.
Thanks to the utilization of the hierarchical architecture, we can naturally apply the pretrained encoder to object detection and semantic segmentation tasks. Empirically, with similar model sizes and FLOPs, MixMAE consistently outperforms BEiT~\cite{beit} and MAE~\cite{mae} on a wide spectrum of downstream tasks, including image classification on iNaturalist~\cite{iNaturalist} and Places~\cite{places}, object detection and instance segmentation on COCO~\cite{coco}, and semantic segmentation on ADE20K~\cite{ade20k}. By abandoning using $\mathrm{[MASK]}$ tokens in the encoder, MixMAE shows much better pretraining efficiency than SimMIM~\cite{simmim} on various hierarchical ViTs~\cite{swin,pvt,chu2021twins}.

\section{Related Works}
\label{sec:related_works}

Inspired by BERT~\cite{bert} for Masked Language Modeling, Masked Image Modeling (MIM) becomes a popular pretext task for visual representation learning~\cite{beit,mae,sit}. MIM aims to reconstruct the masked tokens from a corrupted input.
Current MIM approaches can be divided into two categories by the reconstruction targets. SimMIM~\cite{simmim} points out that raw pixel values of the randomly masked patches are a good reconstruction target and a lightweight prediction head is sufficient for pretraining. Different from SimMIM, MAE~\cite{mae} only takes the visible patches as the input of the encoder. Mask tokens are added in the middle of the encoder and the decoder. Such an asymmetric design greatly reduces the computation overhead of the encoder. To further enhance the feature extraction capability of the encoder, CAE~\cite{cae} separates the encoder and decoder explicitly by adding a feature alignment module in the middle of them. 
Jean-Baptiste et al.~\cite{alayrac2019visual} propose to learn representations by reconstructing original videos from synthetically mixed ones.

Instead of building the reconstruction target manually, using a network to generate the reconstruction target has also been widely applied. In such works, an image tokenizer is used to transform an image into visual tokens. BEiT~\cite{beit} utilizes a pretrained discrete VAE (dVAE)~\cite{dvae,dalle} as the tokenizer. However, the originally used MSE loss in dVAE is insufficient to force the tokenizer to capture high-level semantics. PeCo~\cite{peco} proposed to apply perceptual similarity loss on the training of dVAE can drive the tokenizer to generate better semantic visual tokens, which helps pretraining. Moreover, the tokenizer in BEiT~\cite{beit} needs to be offline pretrained, which limits the model’s adaption ability. To address the problem, iBOT~\cite{ibot} proposed to use an online tokenizer to generate the visual tokens. 

There are also concurrent works that explore using MAE on hierarchical Vision Transformers. UM-MAE~\cite{Li2022ummae} proposed a new masking strategy for adapting MAE to pretrain pyramid-based ViTs (e.g., PVT~\cite{pvt}, Swin~\cite{swin}). GreenMIM~\cite{huang2022green} also adapt MAE on hierarchical architectures. It partitions the local windows into several equal-sized groups and proposes an optimal grouping algorithm to find the optimal group size. Instead of designing specific masking or grouping strategies, we focus on rearranging the inputs and targets. Empirical evaluation shows that MixMAE can obtain better performance with various hierarchical architectures. 

Our MixMAE is also motivated by other mixing-based training techniques~\cite{cutmix,zhang2017mixup,tokenmix}. In contrast to those works which conduct data augmentation for supervised learning, MixMAE can effectively pretrain hierarchical transformers without human annotations. 

\section{Methodology}

In this section, we introduce the proposed MixMAE for learning visual representations via Masked Image Modeling (MIM). We start by briefly revisiting MIM, and then introduce how MixMAE creates training inputs and performs image reconstruction, as well as the hierarchical Transformer architecture.
Finally, we present how to reduce the difficulty of the pretext task to improve the pretraining efficiency.

\subsection{A Revisit of Masked Image Modeling}
Following BERT~\cite{bert}, recent works~\cite{beit,mae,simmim} proposed MIM for learning visual representations. Given an input image $x$, MIM firstly divides the image into non-overlapping image patches $x^p$, following ViT~\cite{vit}. It then samples a random mask $M$ to mask a portion of the image patches, and fills the masked place with a special symbol $\mathrm{[MASK]}$, $\hat{x}^p = x^p \odot M + \mathrm{[MASK]} \odot (1 - M)$, where $\odot$ denotes element-wise multiplication.
The masked image $\hat{x}^p$ is processed by an image encoder to produce the latent representations, and a lightweight decoder (head) is utilized to reconstruct the original image based on the latent representations. The reconstruction target can be chosen as the normalized raw pixel~\cite{mae,simmim} or visual tokens~\cite{beit,peco}. 
MIM computes the mean squared error (MSE) between the reconstructed image patches $y^p$ and the original image patches $x^p$ as the reconstruction loss,  $\mathcal{L}_\mathnormal{rec} = \Vert(y^p - x^p) \odot (1-\mathrm{M})\Vert_2^2$, which is only calculated on masked patches~\cite{mae}. After pretraining, the decoder is discarded and the encoder is used for further finetuning on downstream visual tasks.

\subsection{Mixed and Masked Autoencoder (MixMAE)}
\label{sec:mixmae}

While previous MIM works achieved great progress in self-supervised visual representation pretraining, they usually require a large number of epochs for pretraining. One reason is that they waste much computation on processing the less informative $\mathrm{[MASK]}$ symbols. 
Besides, using the $\mathrm{[MASK]}$ symbol also causes pretraining-finetuning inconsistency as those symbols never appear during finetuning. To tackle the issues, we create mixed images as training inputs from pairs of unlabelled training images, which are generated by mixing two groups of visible tokens from two images, for pretraining. The mixed input is processed by MixMAE to reconstruct original images simultaneously.
For better transferring the learned multi-scale representations to downstream tasks, we utilize the popular Swin Transformer with larger-window size as the encoder of the proposed MixMAE~\cite{swin,swinv2}. Figure~\ref{fig:xmnet} illustrates the proposed framework.
\noindent\textbf{Mixed Training Inputs.} Given two sets of image patches $\{x^p_1, x^p_2\}$ of two random training images, we create a mixed image by filling each spatial location with the corresponding visual token from either $x^p_1$ or $x^p_2$. The mask notation $M$ is slightly abused and we denote $M=1$ as choosing a token from $x^p_1$ and vice versa. The mixed training image $\hat{x}^p_m$ is therefore formulated as:
\begin{equation}
    \hat{x}^p_m = x^p_1 \odot \mathrm{M} + x^p_2 \odot (1-\mathrm{M}).
\end{equation}
MixMAE then takes the mixed image as input for reconstruction during pretraining.
The mixed image no longer consists of the extra $\mathrm{[MASK]}$ symbol and only actual visual tokens, leading to better performances on downstream tasks. The design shares the same principle of MAE~\cite{mae}, but our approach does not disassemble the structure of the 2D image, making it more flexible for adapting to various visual backbones, such as PVT~\cite{pvt} and Swin Transformer~\cite{swin}. We can conduct better pretraining based on various hierarchical vision architectures. We follow common practices to use random masking strategy~\cite{mae,simmim}.

\noindent\textbf{Hierarchical Vision Transformer.} 
For better encoding multi-scale representations, we build the encoder of MixMAE with popular Swin Transformer~\cite{swin} and use larger window size~\cite{swinv2} to encode more context for better reconstruction.

Following Swin Transformer, the input is split into non-overlapping image patches and processed by a linear projection layer.
Then the image patches added with positional embeddings are processed by 4 stages of Transformer blocks to produce hierarchical representations, with a downsampling layer between every two successive stages. However, we do not use the complicated shifted window for information propagation across non-overlapping windows. Instead, we use a relatively large window size (i.e., 14$\times$14), and only conduct global self-attention in stage-3 and -4~\footnote{The feature map resolution is 14$\times$14 / 7$\times$7 for stage-3 / -4. A 14$\times$14 / 7$\times$7 window attention is equivalent to global self-attention. }. The larger window size brings negligible computation overhead, but can better integrate the global context. As we usually use a large masking ratio in MIM, the global context is important for better reconstruction.

We scale the encoder of MixMAE following~\cite{swinv2} with configuration parameters listed bellow: 
\begin{itemize}[leftmargin=*]
    \item Base: $C=(128, 256, 512, 1024)$, $H=(4, 8, 16, 32)$, $B = (2, 2, 18, 2)$,
    \item Large: $C=(192, 384, 768, 1536)$, $H=(6, 12, 24, 48)$, $B = (2, 2, 18, 2)$,
    \item Huge: $C=(352, 704, 1408, 2816)$, $H=(11, 22, 44, 88)$, $B = (2, 2, 18, 2)$,
\end{itemize}
where $C$, $H$, and $B$ denote the channel numbers, numbers of the attention heads, and the numbers of blocks for each stage. The window size is set to $14\times14$ / $7\times7$ for stage-1, -2, and -3 / -4 during pretraining. A linear layer is added between the encoder and the decoder to convert the embedding dimension of the encoder's output to 512. 

\noindent\textbf{Dual Reconstruction.} 
After encoding the mixed input, we \textit{unmix} the token embeddings into two groups according to the binary mask $M$. We then add the $\mathrm{[MASK]}$ tokens to reconstruct the original two images from the two groups with the decoder, which has 8 Transformer blocks with an embedding dimension of 512. The loss is therefore set as
\begin{equation}
\label{eq:loss}
    \mathcal{L}_\mathnormal{rec} = \Vert(y_1^p - x_1^p) \odot (1-\mathrm{M})\Vert_2^2 + \Vert(y_2^p - x_2^p) \odot \mathrm{M}\Vert_2^2,
\end{equation}
where $y_1^p$ and $y_2^p$ are the reconstructed images corresponding to $x_1^p$ and $x_2^p$, respectively. The intuition behind is that as the mixed input contains tokens from two images, we can fully utilize them by reconstructing both images to pretrain the neural network. The computation overhead of reconstructing both images is negligible as the decoder is lightweight.
Our approach demonstrates much higher efficiency than previous works, as to be introduced in Section~\ref{sec:main_results}.

\subsection{Reducing the Difficulty of the Pretext Task}
\label{sec:pretext}

Although the dual reconstruction (Eq. (\ref{eq:loss})) enjoys several benefits, it is a much more challenging optimization problem due to the mixing of image tokens, which causes slow convergence in our preliminary experiments. To reduce the optimization difficulty, we facilitate the dual reconstruction by exploring the following approaches.
\begin{itemize}[leftmargin=*]
    \item \textbf{Mix embedding:} Besides the positional embeddings, we add two mix embeddings to the visual tokens to implicitly differentiate the two mixing groups.
    Each mix embedding is a vector and is shared for tokens from the same image. In practice, we use different mix embeddings for the 4 stages of the encoder and add the embedding at the beginning of each stage.
    \item \textbf{Masked self-attention:} Thanks to the flexibility of the self-attention mechanism, we can also differentiate two mixing images explicitly by masking the self-attention fields. Specifically, for each token, it is only allowed to aggregate information from the tokens belonging to the same image (group). We implement the masked self-attention by reusing the mixing mask $M$ described in Section~\ref{sec:mixmae}. Note that we upsample the mask $M$ by nearest interpolation at different stages to match the resolution of the feature map.
\end{itemize}

\begin{figure*}
    \centering
    \includegraphics[width=1.0\linewidth]{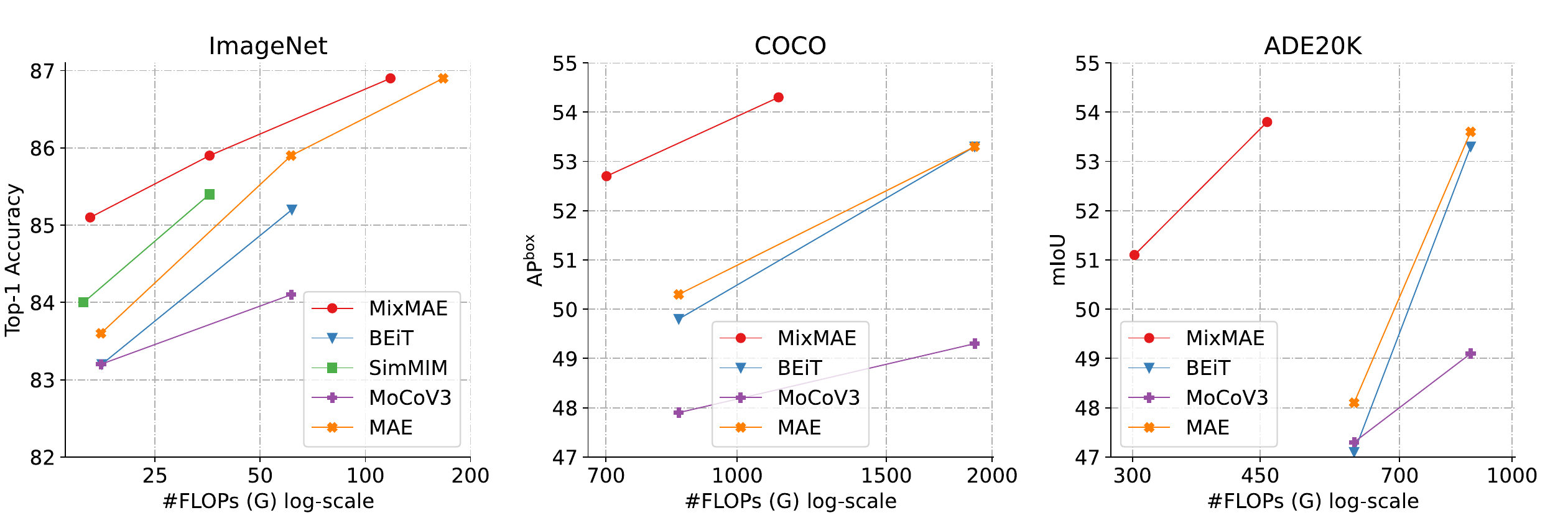}
    \vspace{-1em}
    \caption{Tradeoffs of FLOPs vs. (left) top-1 accuracy on ImageNet-1K, (middle) AP\textsuperscript{box} on COCO, (right) and mIoU on ADE20K. All results are from various self-supervised pretraining methods followed by supervised finetuning. All entries on COCO~\cite{coco} use Mask RCNN~\cite{he2017mask} framework. All entries on ADE20K~\cite{ade20k} use UperNet~\cite{upernet} framework. Note that this comparison confounds differences in architecture and pretraining strategy.}
    \label{fig:ft_acc}
\end{figure*}

Both approaches do not introduce much computation overhead or extra parameters.
The empirical results show that both approaches can help obtain better results. However, the second approach also leads to a faster convergence speed, which is crucial for large-scale pretraining. We use the second approach by default and ablate the design in Section~\ref{sec:ablation}.

\section{Experimental Setup}
\label{sec:setup}

We validate our proposed MixMAE by conducting experiments with pretraining-then-finetuning strategy, following previous practices~\cite{mae,beit}. In particular, we use ImageNet-1K~\cite{imagenet} as the training set for self-supervised pretraining. We then finetune the encoder of MixMAE to downstream tasks, including image classification on ImageNet-1K~\cite{imagenet},  iNaturalist~\cite{iNaturalist}, and Places~\cite{places}, object detection and instance segmentation on COCO~\cite{coco}, and semantic segmentation
on ADE20K~\cite{ade20k}.

\noindent\textbf{Pretraining on ImageNet-1K.} We conduct self-supervised pretraining on ImageNet-1K~\cite{imagenet}. By default, we pretrain for 600 epochs with the input size of $224\times224$. The window size is set as $14\times14$ for the first 3 stages and $7\times7$ for stage-4. 
The patch size of the mask is set to $32\times32$ as our hierarchical encoder eventually downsamples the input to $\frac{1}{32}$ of the input resolution. Following MAE~\cite{mae}, a masking ratio of 75\% is used by default, which is implemented by mixing 4 images. 
We follow all other pretraining hyperparameters of those in MAE~\cite{mae} for a fair comparison. 

\noindent\textbf{Finetuning for image classification.} We conduct supervised finetuning with the pretrained encoder of our MixMAE on image classification tasks, including ImageNet-1K~\cite{imagenet}, Places~\cite{places}, and iNaturalist~\cite{iNaturalist}. We follow previous practices~\cite{beit,mae} and use a layer-wise learning-rate decay strategy~\cite{electra} for finetuning. We sweep the decay rate in \{0.7, 0.75, 0.8\}, and report the best-performing results. We use drop path regularization~\cite{droppath}, and set the drop rate to 0.15/0.2/0.3 for Swin-B/L/H, respectively. We finetune Swin-B/L/H for 100/50/50 epochs following MAE~\cite{mae}. 

\noindent\textbf{Finetuning on COCO.} We perform supervised finetuning on COCO~\cite{coco} for object detection and instance segmentation using the Mask RCNN framework~\cite{he2017mask} with our pretrained encoder as the backbone. We reuse the training setup in MAE~\cite{mae} for a fair comparison.
We change the window size to $16\times16$ for being divisible by the input $1024\times1024$ resolution. Besides, we change the window sizes of the 6th-, 12th-, and 18th-block in stage-3 to 32$\times$32 for cross-window interactions following the previous practice~\cite{vitdet}.

\noindent\textbf{Finetuning on ADE20K.} We perform supervised finetuning on ADE20K~\cite{ade20k} for semantic segmentation. We use the UperNet~\cite{upernet} framework with our pretrained encoder as its backbone. We also change the window size as mentioned above. We reuse the training setup in BEiT~\cite{beit} for a fair comparison.

We include more details about pretraining and finetuning in the Appendix.

\begin{table*}[t]
    \centering
    \resizebox{0.87\linewidth}{!}{
        \begin{tabular}{lccccccc}
            \toprule
            Method & Backbone & FLOPs (G) & Param. (M) & Supervision & Pretrain Epochs & FT & LIN \\
            \midrule
            ViT~\cite{vit} & ViT-B & 17.5 & 86 & RGB & 14 \textsuperscript{$\dagger$} & 79.9 & - \\
            BEiT~\cite{beit} & ViT-B & 17.6 & 87 & DALL-E & 800 & 83.2 & 37.6 \\
            CAE~\cite{cae} & ViT-B & 17.5 & 86 & DALL-E & 800 & 83.6 & 68.6 \\
            MaskFeat~\cite{maskfeat} & ViT-B & 17.5 & 86 & HOG & 300 & 83.6 & - \\
            data2vec~\cite{data2vec} & ViT-B & 17.5 & 86 & Feature & 800 & 84.2 & - \\
            iBOT~\cite{ibot} & ViT-B & 17.5 & 86 & Momentum & 1600 & 84.0 & 79.5 \\
            PeCo~\cite{peco} & ViT-B & 17.5 & 86 & MoCo v3 & 800 & 84.5 & - \\
            MAE~\cite{mae} & ViT-B & 17.5 & 86 & RGB & 1600 & 83.6 & 67.8 \\
            MAE\textsuperscript{$\diamond$}~\cite{mae} & Swin-B/W14 & 16.3 & 88 & RGB & 600 & 84.4 & 61.0 \\
            EsViT\textsuperscript{$\ddagger$}~\cite{esvit} & Swin-B/W14 & 16.3 & 87 & Momentum & 300 & 83.9 & \textbf{81.3} \\
            SimMIM~\cite{simmim} & ViT-B & 17.5 & 86 & RGB & 800 & 83.8 & 56.7 \\
            SimMIM~\cite{simmim} & Swin-B & 15.6 & 88 & RGB & 800 & 84.0 & - \\
            SimMIM\textsuperscript{$\diamond$}~\cite{simmim} & Swin-B/W14 & 16.3 & 88 & RGB & 300 & 84.1 & 20.2 \\
            GreenMIM~\cite{huang2022green} & Swin-B & 15.6 & 88 & RGB & 800 & 83.8 & - \\
            GreenMIM~\cite{huang2022green} & Swin-B/W14 & 16.3 & 88 & RGB & 800 & 84.1 & - \\
            \midrule
            MixMAE & Swin-B & 15.6 & 88 & RGB & 600 & 84.6 & 61.2 \\
            MixMAE & Swin-B/W14 & 16.3 & 88 & RGB & 300 & 84.8 & 63.8 \\
            MixMAE & Swin-B/W14 & 16.3 & 88 & RGB & 600 & \textbf{85.1} & 71.0 \\
            \bottomrule
        \end{tabular}
    }
    \vspace{-1em}
    \caption{Comparison with state-of-the-art MIM methods. All entries are results of base-level models and have comparable model sizes. We report the finetuning accuracy on ImageNet-1K. The FLOPs and Params. are calculated for the encoders. $\dagger$ denotes the number of epochs is based on JFT~\cite{jft300m}. $\ddagger$ results are from~\cite{feature_kd}. $\diamond$ denotes our implementation with the official code.
    FT and LIN denote top-1 accuracy on ImageNet-1K after finetuning and linear probing respectively.}
    \label{tab:sota}
\end{table*}

\section{Main Results}
\label{sec:main_results}

In this section, we compare our MixMAE to prior arts on various visual benchmarks. We present the results on ImageNet-1K~\cite{imagenet} in Section~\ref{sec:imagenet}, and then show the results on the other 6 benchmarks in Section~\ref{sec:transfer}. Note that all the results of MixMAE are obtained by conducting supervised finetuning of the encoder with self-supervised pretraining, without extra intermediate finetuning~\cite{beit}.

\begin{table}[t]
    \centering
    \resizebox{1.0\linewidth}{!}{
        \begin{tabular}{lcccc}
            \toprule
            Pretrain Method & Backbone & \makecell{Pretrain \\ Epochs} &\makecell{ Finetune \\ Epochs} & \makecell{Top-1 \\ Acc.} \\
            \midrule
            Supervised~\cite{swin} & Swin-B & - & 300 & 83.5 \\
            SimMIM~\cite{simmim} & Swin-B & 800 & 100 & 84.0 \\
            GreenMIM~\cite{huang2022green} & Swin-B & 800 & 100 & 83.8 \\
            MixMAE & Swin-B & 600 & 100 & \textbf{84.6} \\
            \midrule
            Supervised~\cite{simmim} & Swin-L & - & 300 & 83.5 \\
            SimMIM~\cite{simmim} & Swin-L & 800 & 100 & 85.4 \\
            GreenMIM~\cite{huang2022green} & Swin-L & 800 & 100 & 85.1 \\
            MixMAE & Swin-L & 600 & 50 & \textbf{85.9} \\
            \midrule
            Supervised~\cite{pvt} & PVT-L & - & 300 & 81.7 \\
            SimMIM\textsuperscript{$\dagger$} & PVT-L & 800 & 100 & 82.0 \\
            MixMAE & PVT-L & 600 & 100 & \textbf{83.4} \\
            \midrule
            Supervised~\cite{chu2021twins} & Twins-SVT-L & - & 300 & 83.7 \\
            SimMIM\textsuperscript{$\dagger$} & Twins-SVT-L & 800 & 100 & 83.3 \\
            GreenMIM~\cite{huang2022green} & Twins-SVT-L & 800 & 100 & \textbf{83.9} \\
            MixMAE & Twins-SVT-L & 600 & 100 & \textbf{83.9} \\
            \bottomrule
        \end{tabular}
    }
    \vspace{-1em}
    \caption{Comparison with state-of-the-art MIM methods using the same encoder. We report the finetuning accuracy on ImageNet-1K. $\dagger$ denotes our implementation with the official code using input size of $224\times224$.}
    \label{tab:encoder}
\end{table}

\subsection{Results on ImageNet-1K}
\label{sec:imagenet}
\noindent\textbf{Comparisons with other MIM approaches.} Table \ref{tab:sota} presents the comparison between MixMAE and state-of-the-art Masked Image Modeling (MIM) works. Our MixMAE can obtain higher accuracy while requiring fewer epochs for pretraining. In particular, we achieve 84.8\% top-1 accuracy with 300 epochs of pretraining, 1.6\% better than BEiT~\cite{beit} with 62.5\% fewer epochs for pretraining. Besides, our MixMAE also enjoys longer pretraining as previous methods~\cite{mae}. Specifically, we obtain strong 85.1\% top-1 accuracy with only 600 epochs of pretraining with Swin-B/W14.

\begin{figure}
    \centering
    \includegraphics[width=0.9\linewidth]{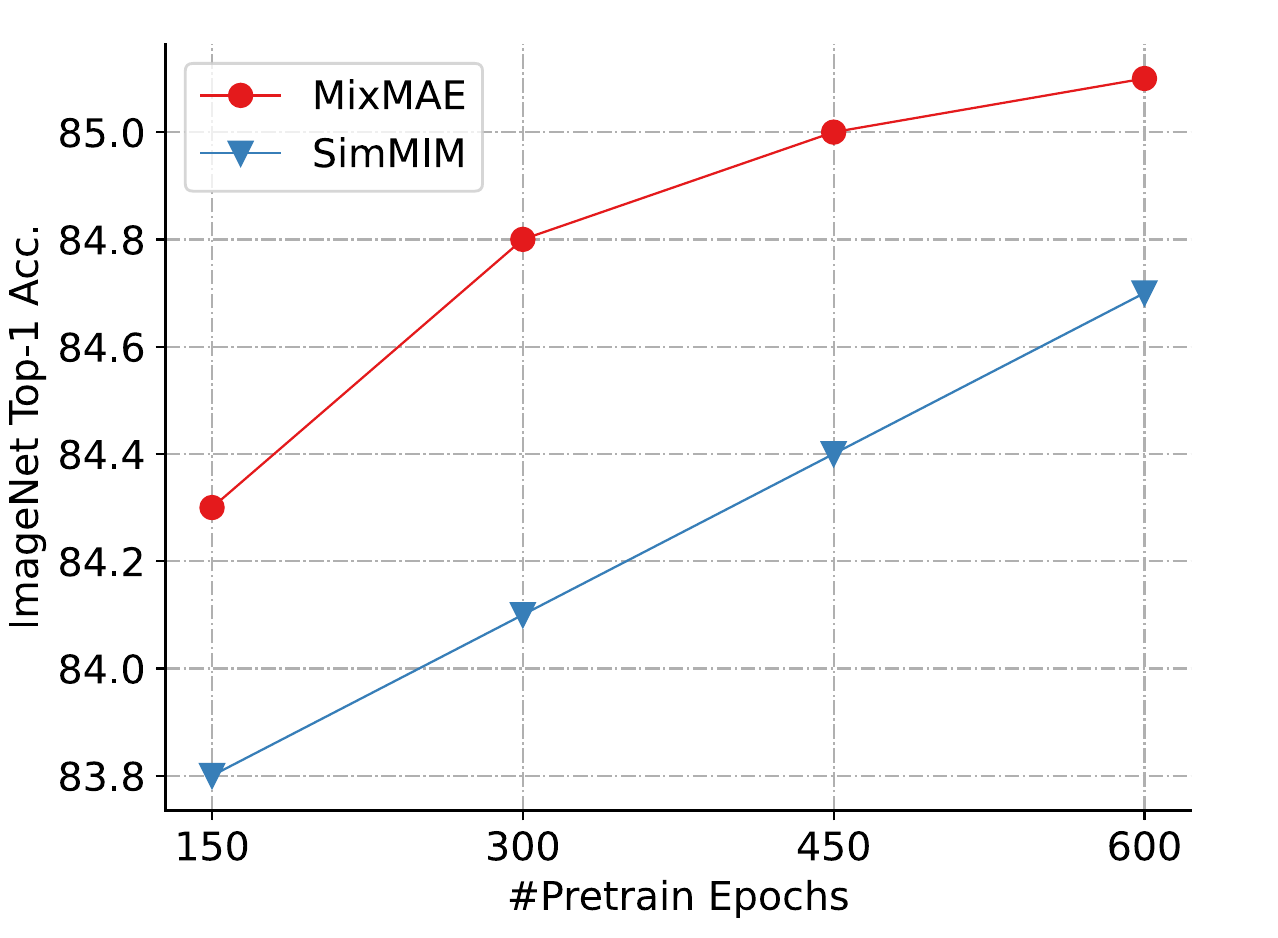}
    \vspace{-1em}
    \caption{Efficiency comparison between MixMAE and SimMIM. We report the finetuning accuracy on ImageNet-1K. The encoder is Swin-B/W14 with input size of $224\times224$.}
    \label{fig:efficiency}
\end{figure}

\begin{table*}[t]
    \centering
    \resizebox{0.95\linewidth}{!}{
    \begin{tabular}{lcc|cccc|ccc}
        \toprule
        \multirow{2}{*}{Method} & \multirow{2}{*}{Backbone} & Pretrain & FLOPs & Params. & \multicolumn{2}{c}{COCO} & FLOPs & Params. & ADE20K \\
        & & Epochs & (G) & (M) & AP\textsuperscript{box} & AP\textsuperscript{mask} & (G) & (M) & mIoU \\
        \midrule
        MoCo v3~\cite{mocov3} & ViT-B & 300 & 853 & 116 & 47.9 & 42.9 & 606 & 164 & 47.3 \\
        BEiT~\cite{beit} & ViT-B & 800 & 853 & 116 & 49.8 & 44.4 & 606 & 164 & 47.1 \\
        MAE~\cite{mae} & ViT-B & 1600 & 853 & 116 & 50.3 & 44.9 & 606 & 164 & 48.1 \\
        iBOT~\cite{ibot} & ViT-B & 1600 & - & - & 51.2 & 44.2 & - & - & 50.0 \\
        EsViT~\cite{simmim} & Swin-B & 300 & - & - & - & - & - & - & 47.3 \\
        SimMIM~\cite{simmim} & Swin-B & 800 & - & - & 52.3 & - & - & - & \textcolor{gray}{52.8\textsuperscript{$\dagger$}} \\
        SimMIM~\textsuperscript{$\diamond$}~\cite{simmim} & Swin-B/W14 & 300 & 701 & 110 & 51.1 & 45.4 & 302 & 122 & 48.9 \\
        GreenMIM~\cite{huang2022green} & Swin-B & 800 & - & - & 50.0 & 44.1 & - & - & - \\
        MixMAE & Swin-B/W14 & 300 & 701 & 110 & 52.3 & 46.4 & 302 & 122 & 49.9  \\
        MixMAE & Swin-B/W14 & 600 & 701 & 110 & \textbf{52.7} & \textbf{47.0} & 302 & 122 & \textbf{51.1}  \\
        \midrule
        MoCo v3~\cite{mocov3} & ViT-L & 300 & 1907 & 339 & 49.3 & 43.9 & 877 & 392 & 49.1 \\
        BEiT~\cite{beit} & ViT-L & 800 & 1907 & 339 & 53.3 & 47.1 & 877 & 392 & 53.3 \\
        MAE~\cite{mae} & ViT-L & 1600 & 1907 & 339 & 53.3 & 47.2 & 877 & 392 & 53.6 \\
        SimMIM~\cite{simmim} & Swin-L & 800 & - & - & 53.8 & - & - & - & \textcolor{gray}{53.5\textsuperscript{$\dagger$}} \\
        MixMAE & Swin-L & 600 & 1119 & 319 & \textbf{54.3} & \textbf{48.2} & 460 & 236 & \textbf{53.8} \\
        \bottomrule
        \end{tabular}
    }
    \vspace{-1em}
    \caption{Comparison with other self-supervised approaches on COCO and ADK20K. We report  AP\textsuperscript{box} and AP\textsuperscript{mask} on COCO, and mIoU on ADE20K. The results of BEiT and MoCo v3 are from MAE~\cite{mae}. The results of EsViT are from ~\cite{feature_kd}. $\dagger$ denotes using supervised finetuning on ImageNet. $\diamond$ denotes our implementation with the official code.}
    \label{tab:det}
\end{table*}

\begin{table*}[t]
    \centering
    \resizebox{0.95\linewidth}{!}{
    \begin{tabular}{lccccccccc}
        \toprule
        Method & Backbone & FLOPs (G) & Params. (M)  & INat2018 & INat2019 & Places205 & Places365 & Average \\
        \midrule
        DINO~\cite{dino} & ViT-B & 17.5 & 86 & 72.6 & 78.2 & - & - & - \\
        MAE~\cite{mae} & ViT-B & 17.5 & 86 & 75.4 & 80.5 & 63.9 & 57.9 & 69.4 \\
        MixMAE & Swin-B/W14 & 16.3 & 88 & \textbf{78.2} & \textbf{83.3} & \textbf{68.6} & \textbf{59.0} & \textbf{72.3}  \\
        \midrule
        MAE~\cite{mae} & ViT-L & 61.3 & 304 & 80.1 & 83.4 & 65.8 & 59.4 & 72.1 \\
        MixMAE & Swin-L & 35.8 & 235 & \textbf{80.6} & \textbf{84.4} & \textbf{69.3} & \textbf{59.6} & \textbf{73.5} \\
        \bottomrule
        \end{tabular}
    }
    \vspace{-1em}
    \caption{Comparison with other self-supervised approaches on classification tasks. We report the top-1 accuracy and average accuracy of all datasets. 
    We also report the average accuracy over the 4 datasets.
    }
    \label{tab:cls}
\end{table*}

While previous works design various reconstruction targets to speed up the pretraining process~\cite{beit,maskfeat}, our MixMAE reconstructs simply normalized pixels~\cite{mae} and demonstrates strong pretraining efficiency. Compared to MaskFeat~\cite{maskfeat}, our MixMAE obtains +1.2\% better accuracy with the same pretraining epochs. PeCo~\cite{peco} proposed to reconstruct the perceptual codebook from a pretrained MoCo v3 network~\cite{mocov3} and can achieve better performance to some extent. In comparison, our MixMAE obtains even better performance (+0.6\%) than PeCo with fewer pretraining epochs (-200). 
The superior performance of MixMAE comes from our mixed pretraining as well as the hierarchical Vision Transformer. However, SimMIM~\cite{simmim} also utilizes a hierarchical Swin Transformer~\cite{swin}, but its performance is worse (-1.1\%) than MixMAE with more pretraining epochs (+200). We conduct a thorough comparison with SimMIM by using the same encoder Swin-B/W14 and show the results in Figure~\ref{fig:efficiency}. Our proposed MixMAE shows much better pretraining efficiency than SimMIM. We list the training time in the Appendix.

We test with scaling MixMAE up to 600M parameters, as shown in Figure~\ref{fig:ft_acc} (Left). MixMAE has better FLOPs vs. accuracy tradeoff than other approaches. In particular, MixMAE pretrained Swin-L and -H achieve 85.9\% and 86.9\% top-1 accuracy, respectively, being comparable with MAE-L (85.9\%) and -H (86.9\%) but requiring fewer FLOPs for inference (-40\% for -L and -30\% for -H).

\noindent\textbf{Integrating MixMAE to other backbones.} While previous works~\cite{mae,maskfeat} may be restricted to a specific architecture, our proposed MixMAE can generalize to various visual backbones, including Swin Transformer~\cite{swin}, Twins~\cite{chu2021twins}, and PVT~\cite{pvt}.
We conduct a thorough comparison with other MIM approaches with fixed encoders. As shown in Table~\ref{tab:encoder}, MixMAE consumes the same or fewer epochs for pretraining but obtains consistently better performance on hierarchical ViTs. In particular, our MixMAE achieves 84.6\% top-1 accuracy with Swin-B, +0.6\% better than SimMIM~\cite{simmim} while requiring 200 fewer epochs for pretraining. With Swin-L, our MixMAE obtains 85.8\% top-1 accuracy with 600 epochs of pretraining and 50 epochs of finetuning, showing higher efficiency than SimMIM. Besides, our MixMAE achieves 83.2\% top-1 accuracy with PVT-L~\cite{pvt}, improving the supervised baseline by a non-trivial margin.

\subsection{Results of Transferring to Downstream Tasks}
\label{sec:transfer}
To further demonstrate the effectiveness of the visual representations learned by MixMAE, we transfer MixMAE to various visual benchmarks with settings described in Section~\ref{sec:setup}.

\noindent\textbf{Object detection and instance segmentation.} We show the results on COCO~\cite{coco} in Table~\ref{tab:det}. By pretraining for 600 epochs on ImageNet-1K, we achieve 52.7 AP\textsuperscript{box} and 47.0 AP\textsuperscript{mask} with Swin-B, surpassing previous self-supervised approaches with fewer FLOPs and parameters. Compared to BEiT~\cite{beit}, our MixMAE obtains higher AP\textsuperscript{box} (+2.9) and AP\textsuperscript{mask} (+2.6) with less pretraining epochs (-200). Note that our hierarchical backbone can naturally be transferred to object detection without re-designing~\cite{vitdet} 
network architectures such as FPN~\cite{fpn}.

Our MixMAE can also scale up to larger models in object detection task and obtains better performance. As shown in Table~\ref{tab:det}, we achieve 54.3 AP\textsuperscript{box} (48.2 AP\textsuperscript{mask}) with Swin-L, +1.0 (+1.1) better than BEiT while requiring 200 fewer epochs for pretraining and 41\% fewer FLOPs for inference. We further evaluate the tradeoff of FLOPs vs. AP\textsuperscript{box} in Figure~\ref{fig:ft_acc} (Middle). We also found that MixMAE outperforms other approaches by large margins.

\noindent\textbf{Semantic segmentation.} Table~\ref{tab:det} also presents the results of MixMAE on ADE20K~\cite{ade20k}. We compare its Mean Intersection over Union (mIoU) on ADE20K with other self-supervised approaches. Our pretrained Swin-B achieves 51.1 mIoU, +4.0 better than BEiT while requiring only half of FLOPs for inference. 
Besides, we obtain 53.8 mIoU by scaling up the model to Swin-L. Thanks to the hierarchical design, our MixMAE consumes much fewer FLOPs for inference compared to other approaches with plain ViT. 
In Figure~\ref{fig:ft_acc} (Right), our MixMAE outperforms other approaches by large margins.

\noindent\textbf{Image classification.} We further transfer MixMAE to other 4 classification datasets and show the results in Table~\ref{tab:cls}. These datasets are challenging as the accuracies are relatively low, e.g., 57.9\% top-1 accuracy on Places365~\cite{places} for MAE-B~\cite{mae}. However, our MixMAE can still outperform previous self-supervised approaches. In particular, we achieve an average +2.9\% performance gain compared to MAE-B with Swin-B. Besides, our pretrained Swin-L has an average +1.4\% performance gain over MAE-L while requiring only 58\% FLOPs for inference.

\section{Ablation Studies}
\label{sec:ablation}

In this section, we ablate the key designs of MixMAE and report the transferring results of each ablation. Unless otherwise specified, we pretrain Swin-B/W14 for 300 epochs with a masking ratio of 50\%. By default, we report the top-1 accuracy on ImageNet-1K~\cite{imagenet} and mIoU on ADE20K~\cite{ade20k}. We also show the results on COCO~\cite{coco} in Appendix.

\begin{figure}
    \centering
    \includegraphics[width=0.95\linewidth]{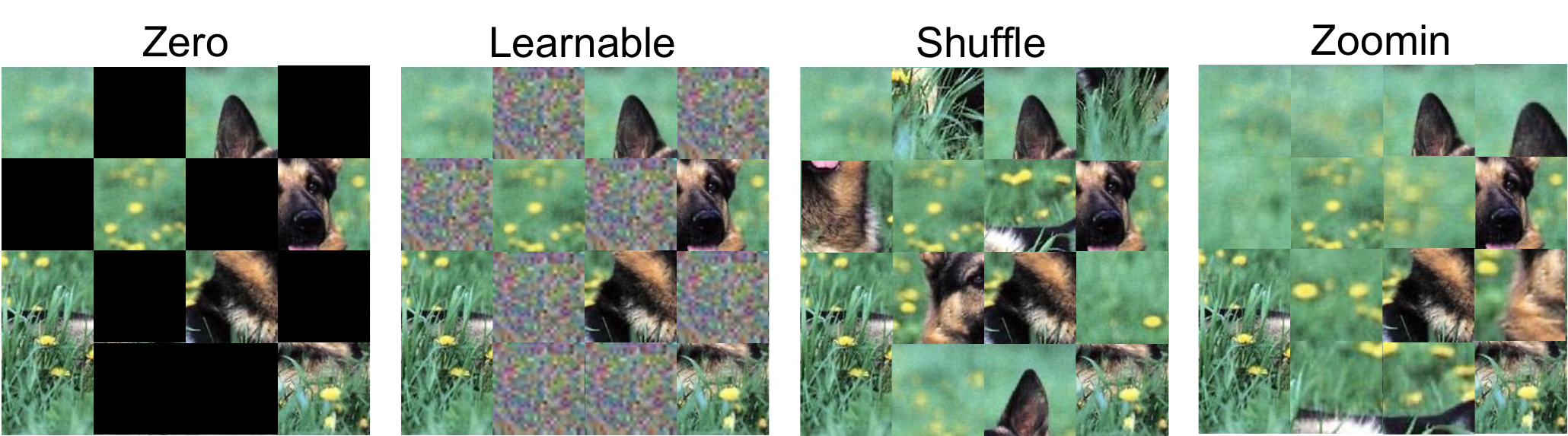}
    \vspace{-1em}
    \caption{Examples images for different filling contents.}
    \label{fig:aba:input}
    \vspace{-1em}
\end{figure}

\begin{table}[t]
    \begin{minipage}[t]{0.47\linewidth}
        \centering
        \resizebox{0.9\textwidth}{!}{
            \begin{tabular}{l|cc}
                \toprule
                Type & Top-1 Acc. & mIoU \\
                \midrule
                \rowcolor{mygray}
                Mix & \textbf{84.6} & \textbf{49.9} \\
                Zero & 84.1 & 48.0 \\
                Learnable & 84.1 & 48.9 \\
                Shuffle & 82.6 & 43.0 \\
                Zoomin & 83.5 & 44.9 \\
                \bottomrule
            \end{tabular}
        }
        \vspace{-10pt}
        \caption{Filling content.}
        \label{tab:aba:input}%
        
        \vspace{10pt}
        
        \centering
        \resizebox{0.9\textwidth}{!}{
            \begin{tabular}{c|cc}
                \toprule
                \# Epochs & Top-1 Acc. & mIoU \\
                \midrule
                300 & 84.6 & 49.9 \\
                600 & \textbf{85.1} & 50.3 \\
                \rowcolor{mygray}
                900 & \textbf{85.1} & \textbf{51.0} \\
                \bottomrule
            \end{tabular}
        }
        \vspace{-10pt}
        \caption{Pretraining epochs.}
        \label{tab:aba:epoch}%
    \end{minipage}
    \hfill
    \begin{minipage}[t]{0.5\linewidth}
        
        \centering
        \resizebox{1.0\textwidth}{!}{
            \begin{tabular}{l|cc}
                \toprule
                \# Images (ratio) & Top-1 Acc. & mIoU \\
                \midrule
                2 (0.5) & 84.6 & 49.9 \\
                2 w/ $\mathrm{[M]}$ (0.75) & 84.4 & 49.0 \\
                3 (0.67) & 84.7	& 49.9 \\
                \rowcolor{mygray}
                4 (0.75) & \textbf{84.8} & \textbf{49.9} \\
                5 (0.8) & 84.5 & 49.5 \\
                \bottomrule
            \end{tabular}
        }
        \vspace{-10pt}
        \caption{Number of mixing images.}
        \label{tab:aba:mask}%
        
        \vspace{2pt}
        
        \centering
        \resizebox{0.8\textwidth}{!}{
            \begin{tabular}{c|cc}
                \toprule
                Dual & Top-1 Acc. & mIoU \\
                \midrule
                \rowcolor{mygray}
                \ding{51} & \textbf{84.6} & \textbf{49.9} \\
                \ding{55} & 84.0 & 47.3 \\
                \bottomrule
            \end{tabular}
        }
        \vspace{-10pt}
        \caption{Dual reconstruction.}
        \label{tab:aba:dual}%
        
    \end{minipage}
\end{table}

\begin{table}[t]
    \centering
    \resizebox{0.8\linewidth}{!}{
    \begin{tabular}{lccc}
        \toprule
        Approach & 300 & 600 & 900 \\
        \midrule
        Ours w/o unmixing & 84.4 & 84.4 & 84.4 \\
        Ours w/ mix embedding & 84.4 & 84.6 & 84.8 \\
        Ours w/ masked self-attention & 84.6 & \textbf{85.1} & \textbf{85.1} \\
        \bottomrule
    \end{tabular}
    }
    \vspace{-1em}
    \caption{Ablation on reducing the difficulty of the pretext task. We report the top-1 accuracy on ImageNet-1K for each approach with different pretraining epochs. Ours w/o unmixing denotes that we do not reduce the difficulty of the pretext task.
    Details of the other two approaches are described in Section~\ref{sec:pretext}. }
    \label{tab:aba:pretext}%
\end{table}

\noindent\textbf{Content to filling.} While MixMAE default fills the masked tokens of one image with visible tokens from another image, we also explore more design choices. Specifically, we try to fill the masked tokens with the following contents.
\begin{itemize}[leftmargin=*]
    \item \textbf{Zero:} Filling the masked tokens with zeros. This approach causes serious mismatches between the masked tokens and the visible tokens.
    \item \textbf{Learnable:} Following previous works~\cite{beit,simmim}, we fill the masked tokens with a shared learnable token. The difference between the zero approach is that learnable tokens can be adapted to visible tokens to match the distribution of the training set.
    \item \textbf{Shuffle:} We randomly shuffle the masked tokens, and then fill the masked locations with the shuffled tokens. We note that this approach is similar to solving jigsaw puzzles~\cite{jigsaw} with the difference that we need to fully regress the pixels.
    \item \textbf{Zoomin:} We zoom in the original image and randomly crop an image patch with the size of the original image. We then fill the masked tokens with tokens from the cropped image. This approach also provides masking tokens that are similar to visible ones but is harder than the shuffle approach.
\end{itemize}
We visualize the four approaches in Figure~\ref{fig:aba:input}. We compare the performances of the four approaches in Table~\ref{tab:aba:input}. Our default choice Mix performs best in terms of accuracy on ImageNet-1K and mIoU on ADE20K. We find the learnable approach has a similar performance on ImageNet-1K but better performance on ADE20K compared to Zero. We hypothesize that the training-finetuning inconsistency has a larger impact on tasks without a lot of labeled images. The shuffle and zoomin approaches perform much worse than other approaches. Those two strategies cause easier pretext task and have lower pretraining loss. However, the learned representation quality is lower.

\noindent\textbf{Dual reconstruction.} We ablate the proposed dual reconstruction in Table~\ref{tab:aba:dual}. We find that dual reconstruction greatly boosts the performance on downstream tasks. The performance gap is larger on ADE20K, where we observe +2.6 mIoU with the dual reconstruction. Note that the computation overhead of dual reconstruction is negligible as the decoder is lightweight.

\noindent\textbf{Masking ratio.} Our MixMAE implements different masking ratios by mixing more images at inputs. In addition, we also experiment to add $\mathrm{[MASK]}$ tokens for a higher masking ratio. We ablate the masking ratios in Table~\ref{tab:aba:mask}. We find that using a masking ratio 75\% by mixing 4 images performs best. In contrast, adding $\mathrm{[MASK]}$ tokens for a 75\% masking ratio has worse performance, demonstrating the effectiveness of the proposed mixing approach.

\noindent\textbf{Pretraining epochs.} Thanks to the dual reconstruction, our MixMAE can achieve strong performance with few pretraining epochs. We ablate the pretraining epochs in Table~\ref{tab:aba:epoch}. We find that the mIoU on ADE20K can be further improved with more pretraining epochs. We achieve 51.0 mIoU with 900 epochs of pretraining. In contrast, the accuracy on ImageNet-1K does not improve after 600 epochs. It might be because the finetuning on ImageNet-1K is more adequate. 

\noindent\textbf{Reducing the difficulty.}
As stated in Section~\ref{sec:pretext}, directly performing reconstruction with the mixed input is a much more challenging optimization problem. Hence, we provide two practical approaches to reduce the difficulty. We ablate the design in Table~\ref{tab:aba:pretext}. We note that all the approaches do not bring nonnegligible FLOPs or parameters. We find that the performance of the approach without unmixing is worst even when trained for more epochs. In contrast, using mix embedding alleviates the problem and improves its performance with longer pretraining. 
However, using masked self-attention in our final solution is much more efficient, and has better performance.

\section{Discussion and Conclusion}
\label{sec:conclusion}
This paper proposes Mixed and Masked AutoEncoder (MixMAE) for efficient visual representation learning. Our MixMAE uses a mixed input created by mixing two (or more) images with random masks, and applies dual reconstruction to recover the original two (or more) images from the mixed hidden representations. We further explore using Swin Transformer with a larger window size for efficient representation learning. Empirical results on 7 visual benchmarks demonstrate MixMAE can learn high-quality visual representations efficiently and has better FLOPs / performance tradeoff than previous MIM works.
While this paper focuses on the vision field, we hope our work will inspire future works in other modalities, such as text and audio. 

\paragraph{Acknowledgement}
This project is funded in part by National Key R\&D Program of China Project 2022ZD0161100,  by the Centre for Perceptual and Interactive Intelligence (CPII) Ltd under the Innovation and Technology Commission (ITC)'s InnoHK, by General Research Fund of Hong Kong RGC Project 14204021. Hongsheng Li is a PI of CPII under the InnoHK.

{\small
\bibliographystyle{ieee_fullname}
\bibliography{egbib}
}
\appendix

\section{Training Details}

\subsection{Hyperparameters of Pretraining and Finetuning}
\label{sec:appdendix:hp}

We include details about the hyperparameters for reimplementation. 

\noindent\textbf{Pretraining.} The default setting is in Table~\ref{tab:pretraing_setting}. We use xavier\_uniform~\cite{xavier} to initialize all Transformer blocks following original ViT~\cite{vit}. We by default use batch size of 1024 and scale the learning rate with linear rule~\cite{goyal2017accurate}: lr=base\_lr $\times$ batch\_size / 256.

\begin{table}[h]
    \centering
    \resizebox{0.45\textwidth}{!}{
    \begin{tabular}{l|c}
        \toprule
        config & value \\
        \midrule
        optimizer & AdamW \cite{adamw} \\
        base learning rate & $1.5\times10^{-4}$ \\
        weight decay & 0.05 \\
        optimizer momentum & $\beta_1,\beta_2$=0.9,0.95~\cite{igpt}  \\
        learning rate schedule & cosine decay~\cite{cosine} \\
        warmup epochs & 40 \\
        augmentation & RandomResizedCrop \\
        \bottomrule
    \end{tabular}
    }
    \caption{Pretraining on ImageNet-1K.}
    \label{tab:pretraing_setting}
\end{table}

\noindent\textbf{Finetuning on ImageNet-1K.} The default setting is in Table~\ref{tab:ft_1k}. We use layer-wise learning rate decay following~\cite{beit,electra}. The decay ratio is swept in \{0.7, 0.75, 0.8\}, and we find 0.7 performs best. Following pretraining, the learning rate is scaled with linear rule: lr=base\_lr $\times$ batch\_size / 256.

\begin{table}[h]
    \centering
    \resizebox{0.46\textwidth}{!}{
    \begin{tabular}{l|c}
        \toprule
        config & value \\
        \midrule
        optimizer & AdamW \\
        base learning rate & $5\times10^{-4}$ \\
        layer-wise lr decay~\cite{beit,electra} & 0.7 \\
        batch size & 1024 \\
        weight decay & 0.05 \\
        optimizer momentum & $\beta_1,\beta_2$=0.9,0.999  \\
        learning rate schedule & cosine decay \\
        warmup epochs & 5 \\
        training epochs & 100 (B), 50 (L/H) \\
        augmentation & RandAug(9, 0.5) \cite{randaug} \\
        LabelSmooth \cite{inception} & 0.1 \\
        Mixup \cite{zhang2017mixup} & 0.8 \\
        CutMix \cite{yun2019cutmix} & 1.0 \\
        drop path \cite{droppath} & 0.15 (B), 0.2 (L), 0.3 (H) \\
        \bottomrule
    \end{tabular}
    }
    \caption{Finetuning on ImageNet-1K.}
    \label{tab:ft_1k}
\end{table}

\noindent\textbf{Finetuning on other classification datasets.} We reuse the setting in Table~\ref{tab:ft_1k}. We adjust the drop path rate for each dataset.

\noindent\textbf{Finetuning on COCO.} We use the Mask RCNN~\cite{he2017mask} framework with the encoder of MixMAE as its backbone. We follow the training setting in~\cite{vitdet,mae}. In particular, we use large-scale jitter~\cite{lsj} augmentation with 1024$\times$1024 resolution and [0.1, 2.0] scale range. We use step learning rate schedule with 0.25 epochs of warmup. We finetune Swin-B/-L for 55/80 epochs. We use a layer-wise learning rate and set the decay ratio to 0.85/0.9 for Swin-B/-L.

\noindent\textbf{Finetuning on ADE20K.} We use the UperNet~\cite{upernet} framework with the encoder of MixMAE as its backbone. We finetune for 16K iterations with a batch size of 16. We use the layer-wise learning rate and set the decay ratio to 0.85/0.9 for Swin-B/-L. We adopt others settings from BEiT~\cite{beit}.

\subsection{Additional Results of Ablation Studies}

\subsubsection{Ablation results on COCO}
\label{appendix:coco}
We show more results of our ablation studies on COCO benchmark in Table~\ref{apptab:aba:input}~\ref{apptab:aba:imgs}~\ref{apptab:aba:epoch}~\ref{apptab:aba:dual}. We find that the performance on the COCO is similar to that on ADE20K.

\begin{table}[t]
    \begin{minipage}[t]{0.45\linewidth}
        \centering
        \resizebox{0.95\textwidth}{!}{
        \begin{tabular}{l|cc}
            \toprule
            Type & AP\textsuperscript{box} & AP\textsuperscript{mask} \\
            \midrule
            Mix & \textbf{51.5} & \textbf{45.9} \\
            Zero & 51.0 & 45.3 \\
            Learnable & 50.9 & 45.1 \\
            Shuffle & 46.5 & 41.6 \\
            Zoomin & 47.9 & 42.6 \\
            \bottomrule
        \end{tabular}
        }
        \caption{Filling content.}
        \label{apptab:aba:input}
    \end{minipage}
    \hfill
    \begin{minipage}[t]{0.55\linewidth}
        \centering
        \resizebox{0.9\textwidth}{!}{
        \begin{tabular}{l|cc}
            \toprule
            \# Images (ratio) & AP\textsuperscript{box} & AP\textsuperscript{mask} \\
            \midrule
            2 (0.5) & 51.5 & 45.9 \\
            2 w/ $\mathrm{[M]}$ (0.75) & 51.2 & 45.4 \\
            3 (0.67) & 51.6	& 45.9 \\
            4 (0.75) & \textbf{52.3} & \textbf{46.4} \\
            5 (0.8) & 51.4 & 45.4 \\
            \bottomrule
        \end{tabular}
        }
        \caption{Number of mixing images.}
        \label{apptab:aba:imgs}
    \end{minipage}
\end{table}

\begin{table}[t]
    \begin{minipage}[t]{0.49\linewidth}
        \centering
        \resizebox{0.85\textwidth}{!}{
        \begin{tabular}{c|cc}
            \toprule
            \# Epochs & AP\textsuperscript{box} & AP\textsuperscript{mask} \\
            \midrule
            300 & 51.5 & 45.9 \\
            600 & 52.2 & 46.5 \\
            900 & \textbf{52.4} & \textbf{46.7} \\
            \bottomrule
        \end{tabular}
        }
        \vspace{-0.2em}
        \caption{Pretraining epochs.}
        \label{apptab:aba:epoch}%
    \end{minipage}
    \vspace{10pt}
    \begin{minipage}[t]{0.5\linewidth}
        \centering
        \resizebox{0.8\textwidth}{!}{
        \begin{tabular}{l|cc}
            \toprule
            Dual & AP\textsuperscript{box} & AP\textsuperscript{mask} \\
            \midrule
            \ding{51} & \textbf{51.5} & \textbf{45.9} \\
            \ding{55} & 50.0 & 44.4 \\
            \bottomrule
        \end{tabular}
        }
        \caption{Dual reconstruction.}
        \label{apptab:aba:dual}
    \end{minipage}
\end{table}

\begin{table}[t]
    \centering
    \resizebox{0.8\linewidth}{!}{
    \begin{tabular}{ccccc}
        \toprule
            Method & Backbone & \makecell{Pretrain Epochs} & \makecell{Top-1 Acc.} \\
            \midrule
            Supervised & ViT-B & - & 81.8 \\
            MAE & ViT-B & 1600 & 83.6 \\
            BEiT & ViT-B & 800 & 83.2 \\
            MixMAE & ViT-B & 600 & \textbf{83.8} \\
        \bottomrule
    \end{tabular}
    }
    \caption{Performance of MixMAE and other methods on ViT.}
    \label{tab:vit}
\end{table}

\subsubsection{Pretraining Time Comparison}
\label{appendix:time}
We compare the wall-clock time of the pretrain in Table~\ref{tab:time}. The pretrain time is measured on 8 NVIDIA-A100-SXM-80GB GPUs with a total batch size of 1024. 
\begin{table}[h]
    \centering
    \resizebox{0.48\textwidth}{!}{
    \begin{tabular}{lcccc}
        \toprule
        Method & Backbone & \makecell{Pretrain \\ epochs} & \makecell{Pretrain Time \\ (GPU hours)} & \makecell{Top-1 \\ Acc.} \\
        \midrule
        SimMIM~\cite{simmim} & Swin-B & 800 & 116 & 84.0 \\
        MAE~\cite{mae} & ViT-B & 1600 & 123 & 83.6 \\
        BEiT~\cite{beit} & ViT-B & 800 & 151 & 83.2 \\
        \midrule
        MixMAE & Swin-B & 600 & 85 & 84.6 \\
        MixMAE & Swin-B/W14 & 300 & 64 & 84.8 \\
        MixMAE & Swin-B/W14 & 600 & 127 & 85.1 \\
        \bottomrule
    \end{tabular}
    }
    \caption{Wall-clock time comparison of MIM methods. }
    \label{tab:time}
\end{table}

\subsubsection{Performance on ViT.}
We show the results on ViT~\cite{vit} in Table~\ref{tab:vit}.

\begin{figure}[t]
    \centering
    \includegraphics[width=0.97\linewidth]{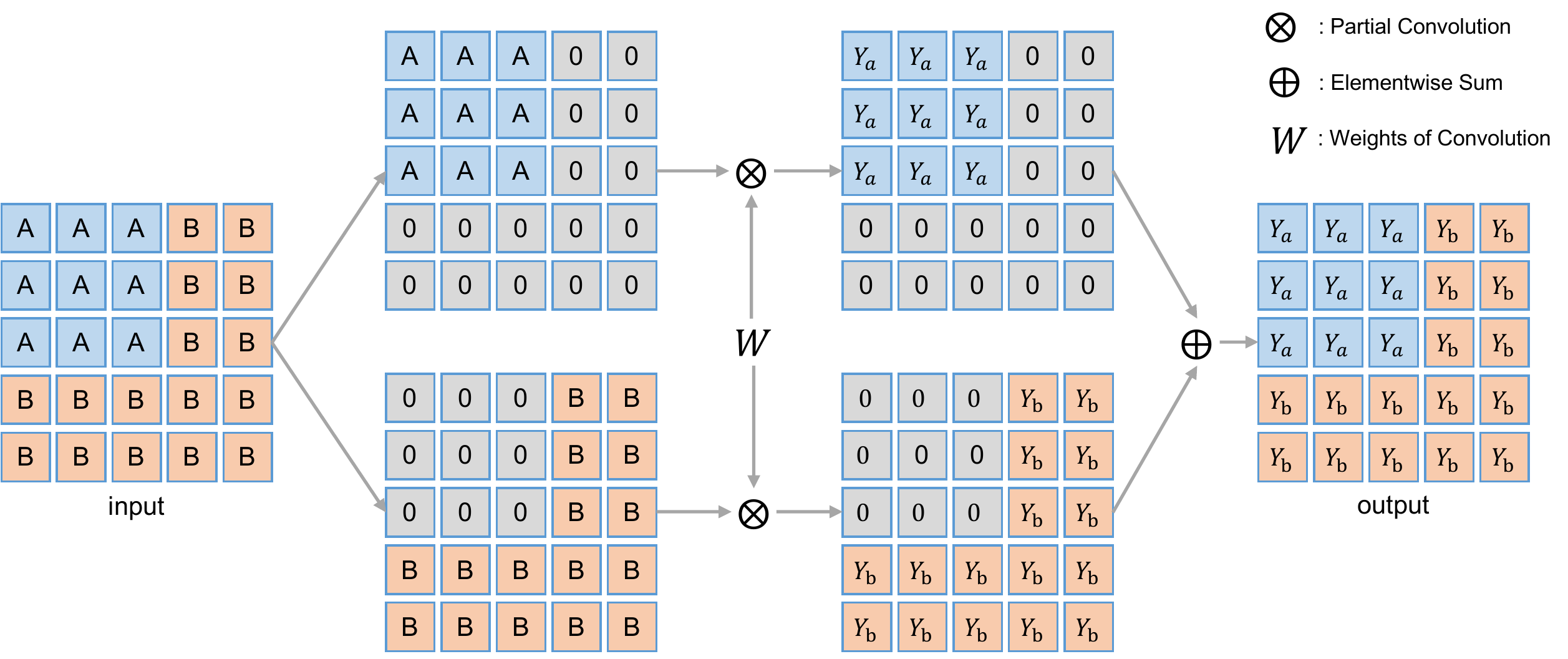}
    \caption{Mixed convolution.}
    \label{fig:mixconv}
\end{figure}

\begin{table}[t]
    \centering
    \resizebox{1.0\linewidth}{!}{
    \begin{tabular}{lcclc}
        \toprule
        Method & Backbone & Input Size & Pretrain Data & Top-1 Acc. \\
        \midrule
        BiT-S~\cite{bit} & Res50x3 & $448\times448$ & ImageNet-1K & 80.0 \\
        BiT-M~\cite{bit} & Res50x3 & $448\times448$ & ImageNet-21K & 84.0 \\
        MixMAE & Res50x3 & $224\times224$ & ImageNet-1K (w/o labels) & 81.8 \\
        \midrule
        BiT-S~\cite{bit} & Res101x3 & $448\times448$ & ImageNet-1K & 80.3 \\
        BiT-M~\cite{bit} & Res101x3 & $448\times448$ & ImageNet-21K & 84.3 \\
        MixMAE & Res101x3 & $224\times224$ & ImageNet-1K (w/o labels) & 82.6 \\
        \bottomrule
    \end{tabular}
    }
    \caption{\textbf{Results on ConvNets.} All results of MixMAE are obtained by pretraining for 300 epochs and finetuning for 100 epochs on ImageNet-1K. We report the top-1 accuracy on ImageNet-1K. }
    \label{tab:convresults}
\end{table}

\subsection{Extend to ConvNets}
While our MixMAE uses a hierarchical Transformer as the encoder, we also explore popular ConvNets. In particular, we use ResNet50x3 and ResNet101x3 as the encoder and compare the finetuning results on ImageNet-1K with BiT~\cite{bit}. To reduce the difficulty of the pretext task, we extend the idea of partial convoluation~\cite{partialconv} and propose a \textit{mixed} version, as illustrated in Figure~\ref{fig:mixconv}.

We compare the results in Table~\ref{tab:convresults}. In particular, our MixMAE outperforms BiT-S by a large margin with half the input size. We note that BiT-M achieves better results by pretraining with 10 $\times$ larger dataset ImageNet-21K. We believe the results of MixMAE can be further improved by using much larger datasets as shown by~\cite{beit}, and we leave it as future work.

\end{document}